\documentclass[10pt,conference,a4paper]{IEEEtran}
\usepackage{cite}
\usepackage[pdftex]{graphicx}
\usepackage{amsmath}
\usepackage{algorithm, algorithmic}
\usepackage{array}

\begin{document}
\title{Latent Model Ensemble with Auto-localization}


\author{\IEEEauthorblockN{Miao Sun\IEEEauthorrefmark{1},
Tony X. Han\IEEEauthorrefmark{1},
Xun Xu\IEEEauthorrefmark{2},
Ming-Chang Liu\IEEEauthorrefmark{2} and 
Ahmad Khodayari-Rostamabad\IEEEauthorrefmark{2} 
\IEEEauthorblockA{\IEEEauthorrefmark{1}Electrical and Computer Engineering\\
University of Missouri,
Columbia, Missouri 65211\\ Email: msqz6@mail.missouri.edu,  hantx@missouri.edu}
\IEEEauthorblockA{\IEEEauthorrefmark{2}Sony Electronics Inc, San Jose, California, 95112\\
Email: Xun.Xu, Ming-Chang.Liu, Ahmad.Khodayari@am.sony.com}
}}

\maketitle

\begin{abstract}
Deep Convolutional Neural Networks (CNN) have exhibited superior performance
in many visual recognition tasks including image classification, object
detection, and scene labeling, due to their large learning capacity and
resistance to overfit. For the image classification task, most of the
current deep CNN-based approaches take the whole size-normalized image as
input and have achieved quite promising results. Compared with the previously
dominating approaches based on feature extraction, pooling, and classification,
the deep CNN-based approaches mainly rely on the learning capability of deep CNN
to achieve superior results: the burden of minimizing intra-class
variation while maximizing inter-class difference is entirely dependent on 
the implicit feature learning component of deep CNN; we rely upon the
implicitly learned filters and pooling component to select the discriminative
regions, which correspond to the activated neurons. However, if the
irrelevant regions constitute a large portion of the image of interest, the
classification performance of the deep CNN, which takes the whole image as input, can be heavily affected. To solve this issue, we propose a novel latent
CNN framework, which treats the most discriminate region as a latent variable. We
can jointly learn the global CNN with the latent CNN to avoid the aforementioned
big irrelevant region issue, and our experimental results show the evident
advantage of the proposed latent CNN over traditional deep CNN:
latent CNN outperforms the state-of-the-art performance of deep CNN on
standard benchmark datasets including the CIFAR-10,
CIFAR-100, MNIST and PASCAL VOC 2007 Classification dataset.  
\end{abstract}


\IEEEpeerreviewmaketitle

\section{Introduction}
\label{sec:intro}

\begin{figure}[!t]
\centering
\includegraphics[width=\linewidth]{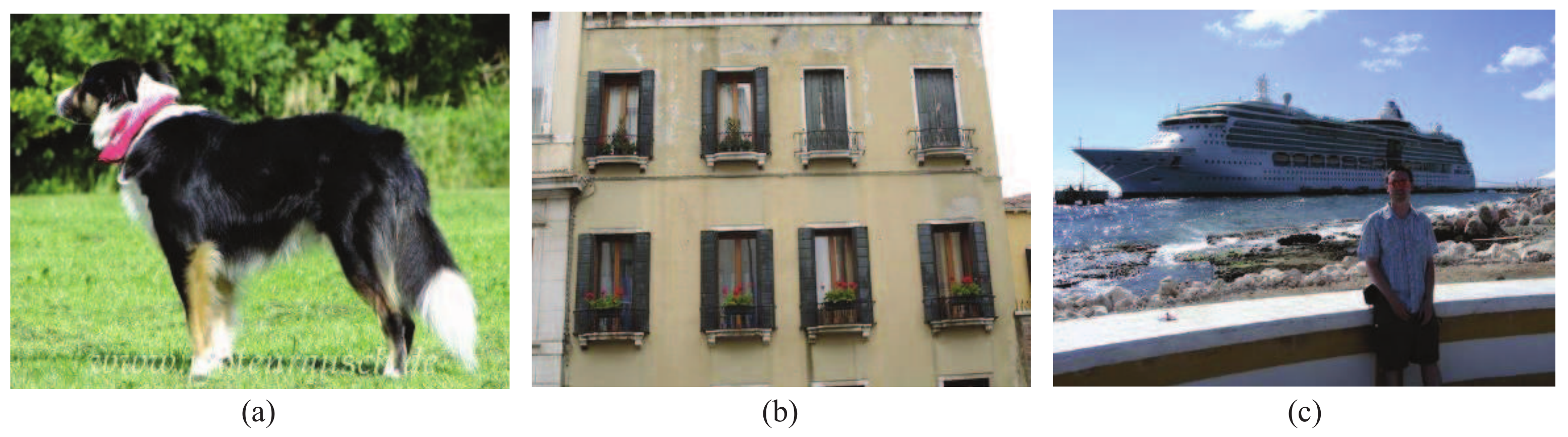}
\caption{Some PASCAL VOC 2007 images. Irrelated regions increase the complexity of CNN learning, which is especially evident in Fig. 1(b) and (c). Figure (a) (b) are labeled as dog and potted plant respectively. Figure (c) has two labels: boat and person.}
\label{fig:idea}
\end{figure}

The last decade of progress on various visual recognition tasks has for the most part been based on the use of SIFT~\cite{Sift} and HOG~\cite{HOG}. The recent success of CNNs is attributed to their ability to learn rich mid-level image representations as opposed to hand-designed low-level features used in other image classification methods. ~\cite{Miao} has demonstrated that deep CNN features are substantially different from and complementary to those traditional features used in object detection.
Searching the parameter space of deep architectures is a difficult task because the training criterion is non-convex and involves many local minima. Nevertheless, deep architecture is capable of automatically learning and fusing rich hierarchical features in an integrated framework. Many techniques, such as Relu~\cite{Alex}, Dropout~\cite{Dropout}, Dropconnect~\cite{Dropconnect}, pre-training~\cite{Pretraining} and data augmentation~\cite{mDNNmnist}, have been proposed to enhance the performance of deep architectures.  Though learning CNN will get into local minima or in a plateau (where due to low curvature the gradients become extremely small), deep convolutional neural networks recently achieved remarkable success in many visual recognition tasks, such as image classification and object detection, fine-grained recognition, and visual instance retrieval~\cite{offtheshelf}.  
For the image classification task, most of the current deep CNN-based approaches take the whole size-normalized image for input. However, it is very likely that the region of interest in the image may just take a small portion of the image of interest, especially for the object classification task of PASCAL VOC dataset. Figure~\ref{fig:idea} shows some images from the PASCAL VOC 2007 classification dataset ~\cite{PASCAL07}. Figure~\ref{fig:idea}(a) is a perfect example for whole image input data, which is centered and occupies a large portion of the image. However, Figure~\ref{fig:idea}(b), which is labeled "potted plant", would make the learning more complicated for CNN's use of the whole image as input. The most challenge is that Figure~\ref{fig:idea}(c) has multiple labels. Currently supervised learning CNN takes (data, label) pairs as input; Hence, multiple labeled images would tend to confuse the CNN model.

In order to reduce the effect of irrelated regions,  we propose a novel framework called latent CNN, which would select the most discriminating region as input for deep CNNs. In this view, latent CNN could also be seen as region-level CNN instead of traditional image-level CNN, which takes the whole size-normalized image as input.

One straightforward way to reduce the effect of local minima is to make full use of multiple CNNs with different random initialization~\cite{Alex}. Given multiple CNNs, people simply use majority voting or average the confidence scores from different CNNs. The second contribution in our paper is that we propose a new combination scheme called Latent Model Ensemble, which outperforms the state-of-the-art performance on CIFAR-10~\cite{Cifar}, CIFAR-100~\cite{Cifar}, MNIST~\cite{Mnist}, and PASCAL VOC 2007 Classification dataset~\cite{PASCAL07}.

In summary, this paper introduces the following contributions: (i) a latent CNN framework, which automatically selects the most discriminate region to reduce the effect of irrelevant regions, (ii) a new combination scheme for multiple CNNs via Latent Model Ensemble,  and (iii) state-of-the-art performance on CIFAR-10~\cite{Cifar}, CIFAR-100~\cite{Cifar}, MNIST~\cite{Mnist} and PASCAL VOC 2007 Classification dataset~\cite{PASCAL07}.

\section{Latent CNN Framework}
Latent CNN contains two key components: deep CNN structures and latent SVM, which we will present in detail in Sections ~\ref{sec:cnn} and ~\ref{sec:svm}, respectively. Then we will discuss the overall latent CNN framework in ~\ref{sec:lcnn}.

\subsection{Deep CNNs}\label{sec:cnn}


AlexNet~\cite{Alex} is a stack of convolutional layers, which are optionally followed by contrast normalization layers and max-pooling layers, and locally-connected layers or fully-connected layers. Transfer learning is adopted when the AlexNet is used for other smaller dataset. For example, the PASCAL VOC 2007 classification dataset~\cite{PASCAL07} has only 5,011 training images, which makes it almost impossible to learn a satisfied deep CNN model.


Our CNN structure trained on ImageNet is similar to ~\cite{Alex}, which has five convolutional layers, two fully connected layers, and one output layer. The max-pooling layers are added to conv1, conv2, and conv5 layer, respectively. The difference is that ~\cite{Alex} normalized all the images into 256 x 256 squares and take 224 x 224 patches as the input for the CNN structure. Our CNN structure, however, does not need to normalize the original images and it takes 128x128 patches. Compared to the AlexNet image translation method, our method may not be label-preserving transformations~\cite{bestPrac, mDNNmnist,highNN, cao}, considering that a much smaller region probably does not satisfy the 50\% overlap rule. But it is not a problem when it is applied to the latent CNN framework, which would select the most discriminative region as the input for CNN learning procedure.

Transfer learning is used for training the CNN structure for the PASCAL VOC datasets. Compared to the CNN model trained from ImageNet, it removes the last 1000-node output layer and add two more fully connected layers. During the training for the PASCAL VOC 2007 classification dataset, all the transferred parameters are fixed first and only the parameters in the last two fully connected layers are updated. Finally, a fine tuning of the whole CNN structure is applied.

%
%

\subsection{Multi-class Latent SVM}\label{sec:svm}

Before formulating the multi-class latent SVM, let's first see the classical SVM~\cite{SVM}:

Assume we are given (label, feature-vector)  pairs of training data $(y_i, \mathbf{x_i}), i=1,...,n$. Linear classification involves the following optimization problem

\begin{equation}
  \min_\mathbf{w} f(\mathbf{w}) \text{, where } f(\mathbf{w}) = \frac{1}{2} \mathbf{w}^T \mathbf{w} + C\sum_{i=1}^{n} \xi(\mathbf{w};\mathbf{x_i}, y_i ),
\label{eq:svm}
\end{equation}

where $\xi(\mathbf{w};\mathbf{x_i}, y_i )$ is the loss function, and C is a  penalty parameter. Common loss functions include

\begin{equation}
  \xi(\mathbf{w};\mathbf{x_i}, y_i ) = max(0, 1-y_i \mathbf{w}^T\mathbf{x_i}) ,
\end{equation}  

\begin{equation}
\label{eq:L2}
  \xi(\mathbf{w};\mathbf{x_i}, y_i ) = max(0, 1-y_i \mathbf{w}^T\mathbf{x_i})^2 .
\end{equation}

In analogy to classical SVMs , the latent SVM can be formulated as
\begin{equation}
\begin{split}
  &\quad \min_\mathbf{w} f(\mathbf{w}) \text{, where } f(\mathbf{w}) = \\
  &\quad \frac{1}{2} tr(\mathbf{w}^T \mathbf{w}) + C\sum_{i=1}^{n} min_{z \in Z(\mathbf{x_i})}(\xi(\mathbf{w};\mathbf{x_i}, y_i , Z)),
\end{split}
\end{equation}

We propose a Stochastic Gradient Descent version of latent SVM with L2 loss function , which is summarized in Algorithm \ref{algm:MLSVM}. Notice that Algorithm \ref{algm:MLSVM} is a multi-class latent SVM. Model parameter matrix $\mathbf{w}$ is  $FL$ x $NC$ matrix, where $FL$ is the feature length for $\Phi(\mathbf{x_i}, z)$, and $NC$ is the number of categories. Label $y(\mathbf{x_i})$ is written into a $NC$ dimensional vector format in order for matrix multiplication.

\begin{algorithm}[htp]
  \caption{\small{SGD version of Multi-class Latent SVM}}
  \label{algm:MLSVM}
  \begin{algorithmic}
    \STATE {\bfseries Input:} Feature vector $\Phi(\mathbf{x_i}, Z)$, $i = 1 ... n$, $Z$ is the latent variable space, label vector $y(\mathbf{x_i})$, epoch numbers $EN$, penalty parameter $C$, learning rate $lr$
    \STATE {\bfseries Input:} Model parameter matrix $\mathbf{w}$
    \STATE $t=1$ , $t$ is the gradient update times
    \FOR{$j=1$ {\bfseries to} $EN$}
      \FOR{$i=1$ {\bfseries to} $n$}
        \STATE 1. The L2 Loss is $\xi(\mathbf{w};\mathbf{x_i}, y_i , z)$,
        \STATE 2. Find the $z_{max}$ which will maximize $\xi$ and the correspond L2 loss is $\xi_{max}$,
        \STATE 3. The objective function loss is $\frac{1}{2} \mathbf{w}^T \mathbf{w} + C\xi_{max}$,
        \STATE 4. The model parameter update gradient is $G = \mathbf{w} - 2 C \Phi(\mathbf{x_i}, z_{max})  \xi_{max}  y(\mathbf{x_i}, z_{max})$,
        \STATE 5. Then we update the model vector $\mathbf{w} = \mathbf{w} - G/(t+T) * lr$, where $T$ is a very large fixed number
        \STATE 6. $t = t + 1$.
      \ENDFOR
    \ENDFOR
  \end{algorithmic}
\end{algorithm}

\subsection{Latent CNN Framework}\label{sec:lcnn}

\begin{figure}[!t]
\centering
\includegraphics[width=\linewidth]{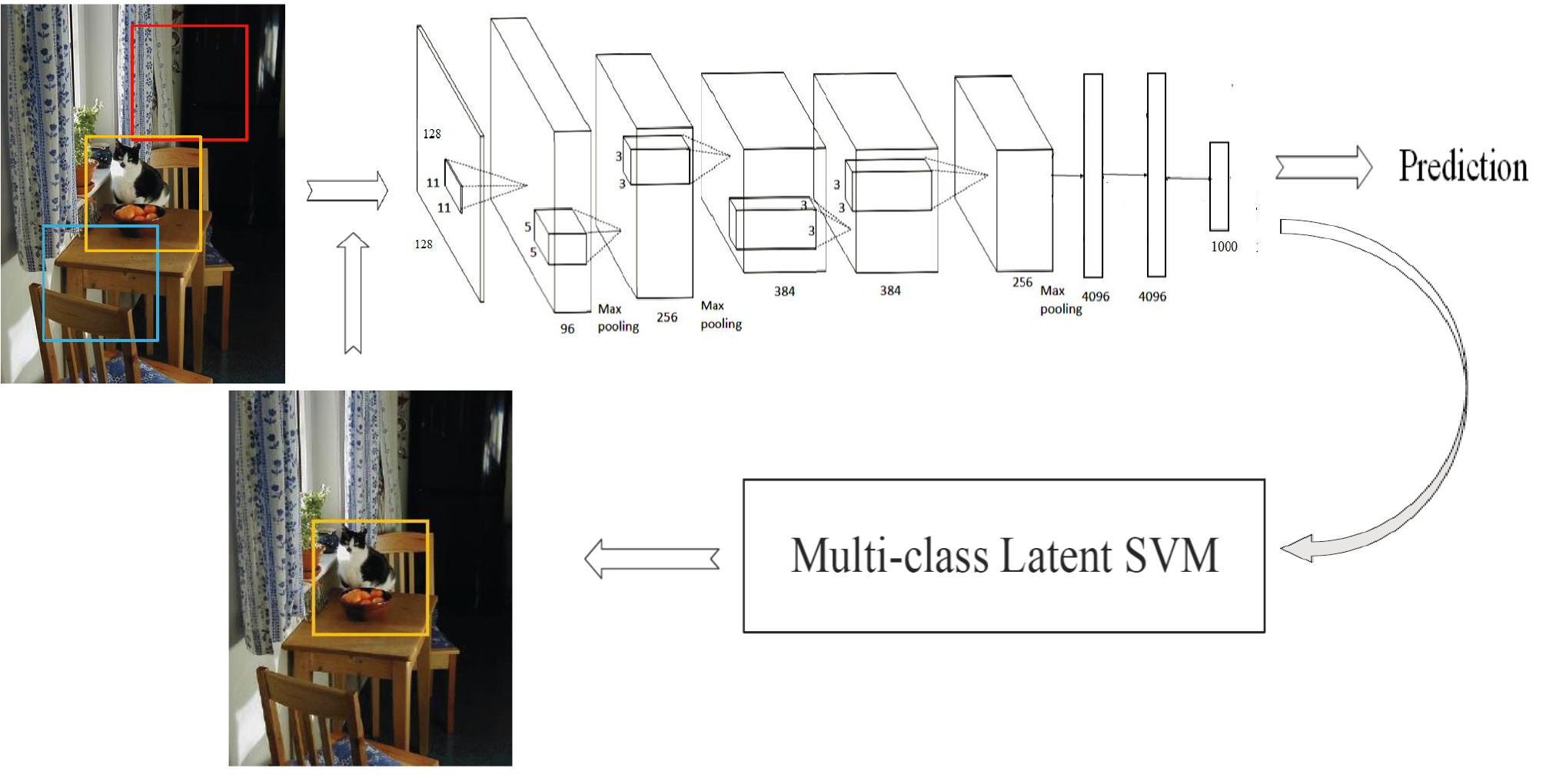}
\caption{Latent CNN framework. The red, yellow and blue rectangular boxes are randomly selected regions. After the deep CNN representation stage and latent SVM selecting stage, only the yellow region is used to update the CNN structure.}
\label{fig:lcnn}
\end{figure}

In Figure~\ref{fig:lcnn}, the latent CNN framework consists of a deep CNN representation stage and latent SVM selecting region stage. Given one image, the deep CNN representation stage treats all the random selected regions as input. Then the latent SVM takes those deep CNN features and selects the most discriminating region. And the selected region is used to update the parameters in the deep CNN structure.

\begin{figure}[!t]
\centering
\includegraphics[width=\linewidth]{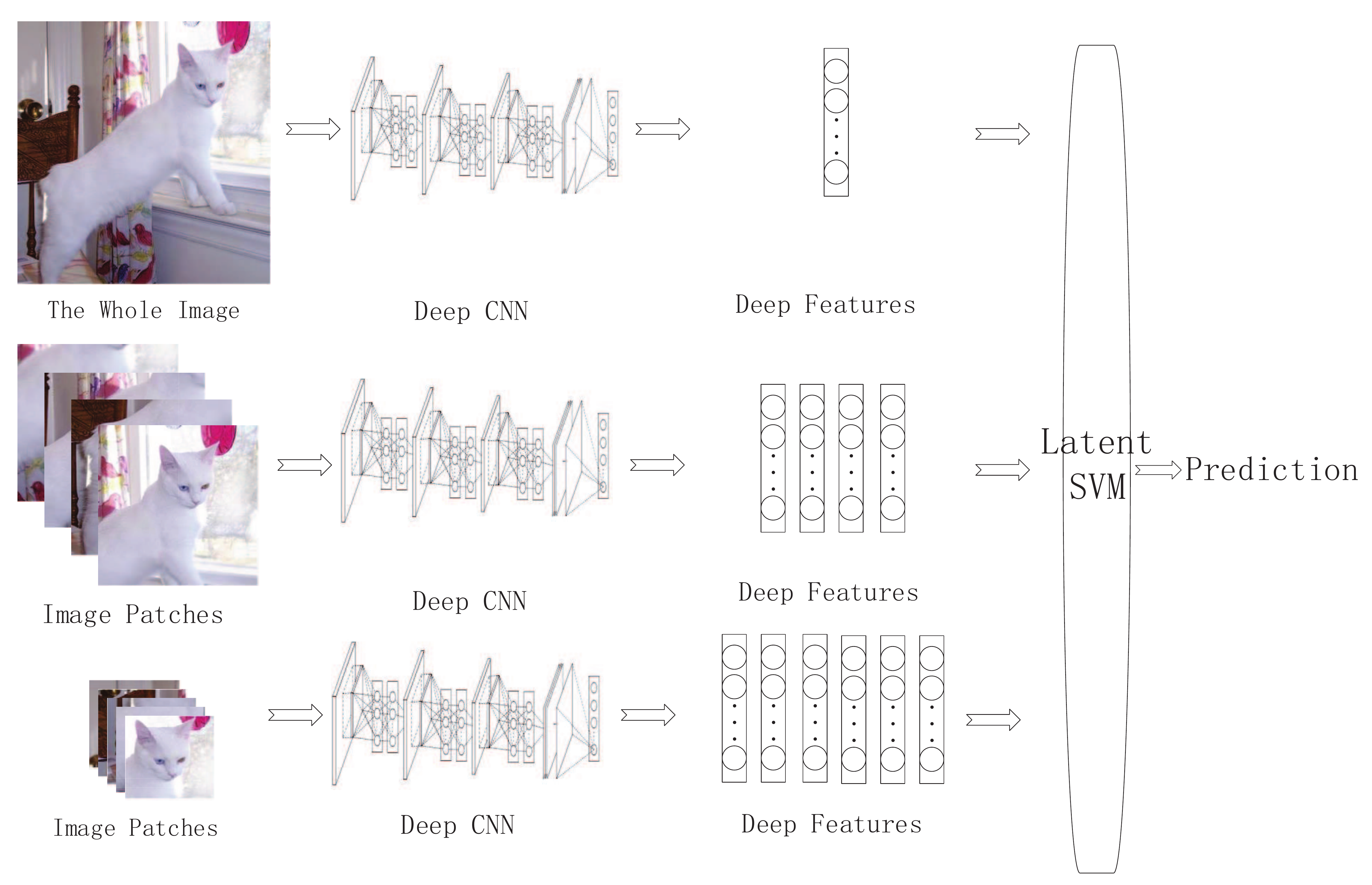}
\caption{Latent Model Ensemble with Deep Features. \footnotesize{The CNN in the first row is trained with the whole image, and the CNN in the second row is trained with the image patches. The latent SVM is used to select the max response from a group of patches, which is used to combine with the deep features in the first row as in Algorithm \ref{algm:MLSVM2}. }}
\label{fig:framework}
\end{figure}

Felzenszwalb et al.~\cite{DPM} generalized SVMs for handling latent variables such as part positions, which is called Latent SVM (LSVM). They also proved that a latent SVM, like a hidden CRF~\cite{CRF}, leads to a non-convex training problem. However, unlike a hidden CRF, a latent SVM is semi-convex and the training problem becomes convex once latent information is specified for the positive training examples.

For a detector with a single component in ~\cite{DPM}, the model is defined by a coarse root filter, several higher resolution part filters and a spatial model for the location of each part relative to the root.  The latent SVM is developed to train the filters and the relationship between them.  

Inspired by~\cite{DPM}, the deep CNN features extracted from the whole image are considered as the features for the 'root' models and those features from part of the image are considered as the features for 'part' models. In this view, our algorithm can be summarized in Algorithm \ref{algm:MLSVM2} and the framework is illustrated in Figure~\ref{fig:framework}.

\subsection{Latent Model Ensemble}\label{sec:lme}
One straightforward way to reduce the effect of local minima is to make full use of multiple CNNs with different random initializations~\cite{Alex}. Given multiple CNNs, people simply use majority voting or average the confidence scores from different CNNs. Latent Model Ensemble is developed based on Latent SVM\cite{DPM}. Algorithm~\ref{algm:MLSVM2} is the detailed description for Latent Model Ensemble. We define the CNN trained on the whole image as root model and the CNN trained on patches of images as part model. The usage of latent SVM in Algorithm~\ref{algm:MLSVM2} has two aspect: one is used for choosing the most discrimating patch, which is used to update the latent CNN framework as shown in Figure~\ref{fig:lcnn}, while the other is for selecting proper weight of root features and part features. 

\begin{algorithm}[!htp]
  \caption{Latent Model Ensemble}
  \label{algm:MLSVM2}
  \begin{algorithmic}
    \STATE {\bfseries Input:} Feature vector $\Phi(\mathbf{x_i}, Z)$, $i = 1 ... n$, $Z$ is the latent variable space, label vector $y(\mathbf{x_i})$, epoch numbers $EN$, penalty parameter $C$, learning rate $lr$
    \STATE {\bfseries Input:} Model parameter matrix $\mathbf{w}$

    \FOR{$j=1$ {\bfseries to} $EN$}
      \FOR{$i=1$ {\bfseries to} $n$}
        \STATE 1. Crop image into patches, and extract deep features from each patch
        \STATE 2. Update the part model according to step1-step5 in the inner loop of Algorithm \ref{algm:MLSVM}
        \STATE 3. Extract deep features from the whole image
        \STATE 4. Update the root model according to Equations (\ref{eq:svm}) and (\ref{eq:L2}).
      \ENDFOR
    \ENDFOR
  \end{algorithmic}

\end{algorithm}


\section{Experiments}

We evaluate our framework based on four benchmark datasets: CIFAR-10 ~\cite{Cifar}, CIFAR-100 ~\cite{Cifar}, MNIST~\cite{Mnist}, and PASCAL VOC 2007 Classification dataset ~\cite{PASCAL07}. To extract deep features, we use Network In Network (NIN)~\cite{NIN} for the first three datasets and AlexNet~\cite{Alex} for the PASCAL dataset. For all the datasets we use 10 part locations, designated as top left, top right, center, down left, down right, and the horizontal flip.

PASCAL VOC2007 datasets have 20 categories and contain 9,963 images. This dataset is divided into train, validation, and test subsets, which contains 2,501, 2,510 and  4,952 images, respectively.  The datasets are extremely challenging since the objects vary significantly in size, view angle, illumination, appearance and pose. Some example images in the PASCAL classification task can be seen in Figure~\ref{fig:idea}. The object in most images are not well-centered, and one image may even contain more than one label.  The naive thought for image classification is using the whole image for both training and testing procedure. However, due to the specification of the PASCAL images, people are allowed to use additional information for the training procedure, that is, bounding boxes which are axis-aligned rectangles specifying the extent of the object visible in the images. When the bounding boxes are used in the training procedure, people often crop the patches from test images for evaluation in order to keep consistent. In order to train a satisfied deep CNN, we pretrain the CNN on an ILSVRC2012 train dataset with 1,000 categories and 1.2 million images. The structure for pretrained CNN is similar to AlexNet, except that only one GPU is used. The preprocessing of the dataset simply entails resizing the images into 256 x 256 x 3 without keeping the aspect ratio and random crop 224 x 224 x 3 for data augmentation. The network is composed of five successive convolutional layers C1, C2, C3, C4, C5 followed by three fully connected layers FC6, FC7, and FC8. The trained CNN achieves an error rate of 19.7\% for top5, which reproduces Alex's~\cite{Alex} performance. To achieve the transfer learning from 1000 category ILSVRC dataset to 20 category PASCAL dataset, we remove the output layer FC8  and add additional two fully connected layers with 2,048 and 21, separately. Notice that we add one additional category called background, so the number of outputs was set as 21; if we use the whole image as training examples, of course, the category number was 20, which is also called image-level classification in Table~\ref{tab:input}. 

For the remaining datasets, another popular convolutional network, called NIN~\cite{NIN}, was adopted, and the network structure was the same as~\cite{NIN}. 

\subsection{Effect of Data Input}

Given a fixed neural network structure, the training data would be the major factor for better optimization solutions. In order to make the training task easier, one common scheme is to apply preprocessing to the training data and the test data, such as demean in~\cite{Alex}. For image classification, especially for PASCAL format dataset where each image has a bounding box corresponding to the image label, we were able to take advantage of the bounding boxes to produce additional training data via the sliding-widow cropping method or segmentation method. 

In Table~\ref{tab:input}, the patch-slide methods are cropping images with sliding window strategy and extracting around 500 square patches from each image using the same method in~\cite{MidCnn}.  In this paper, we propose a new data input format, that is, we make use of image segmentation~\cite{Seg} to create image patches. During the segmentation, we automatically remove patches that are too small or too big aspect ratio. After each image had roughly 1,000 patches related to it, then we label those patches into 21 category with following rules: (1) 70\% of the object pixels should be in the patches, (2) the patch can overlap with no more than one object. In fact, there were too many negative boxes, so we roughly kept 10\% negative boxes. So the total patches for the new CNN were 1,398,722 images, that is, each image has about 282 patches. For the testing procedure, we also use the image segmentation; then the final confidence score was the average of all the scores from all the patches from the corresponding image.

In Table~\ref{tab:input}, image-level input format performs much worse than patch-level format, and the patch-seg format performs best. The last row segTrainSlideTest means we train the CNN model using the segmented patches and use the sliding-window patches for testing, so the last two rows share the same CNN model and the difference is just the input of testing procedure. We believe that the segmented patches are best because it provides well-centered object images for training and testing.

\begin{table*}[t]
  \caption{CNN Classiﬁcation on PASCAL VOC 2007 with different input data(average precision \%)}
  \label{tab:input}
  \vskip 0.15in
  \begin{center}
  \begin{small}
  \begin{sc}
 \setlength{\tabcolsep}{3pt}
 \begin{tabular}{|c||c|c|c|c|c|c|c|c|c|c|c|}
\hline
\;\; &plane&bike&bird& boat& bottle& bus& car& cat& chair& cow
\\\hline
Image-level &83.3&69.7&80.0&73.5&34.9&61.5&80.8&76.2&53.0&68.6
\\\hline
Patch-slide(1 scale) &77.9&75.0&79.4&71.8&27.7&75.2&78.6&81.0&39.0&65.5
\\\hline
Patch-slide(8 scale) &82.2&78.8&81.8&78.6&53.3&77.2&87.6&80.2&\bf63.8&\bf79.9
\\\hline
segTrainSlideTest   &86.1&81.3&86.4&81.1&54.1&78.8&88.8&\bf86.7&58.2&72.5
\\\hline
Patch-seg   &\bf86.9&\bf85.3&\bf87.7&\bf82.9&\bf60.4&\bf79.3&\bf89.8&86.1&63.7&75.8
\\\hline
\;\; &table& dog& horse& motor& person& plant& sheep& sofa& train& tv& mAP
\\\hline
Image-level &62.4&73.9&83.1&67.5&83.0&48.9&73.5&60.4&82.5&60.0&68.8
\\\hline                                                            
Patch-slide(1 scale) &\bf74.3&79.0&\bf86.8&77.6&85.0&48.2&72.9&72.2&88.0&55.2&70.5
\\\hline
Patch-slide(8 scale) &67.7&80.2&84.4&78.4&92.6&62.8&\bf80.9&73.2&85.6&78.2&77.4
\\\hline                                                            
segTrainSlideTest  &72.4&84.1&80.3&80.2&93.8&66.1&78.9&69.0&88.7&79.6&78.4
\\\hline
Patch-seg   &73.4&\bf85.6&83.6&\bf84.6&\bf94.1&\bf68.8&80.1&\bf74.1&\bf89.6&\bf79.8&\bf80.6
\\\hline

\end{tabular}
\end{sc}
\end{small}
\end{center}
\vskip -0.1in
\end{table*}

\subsection{Latent CNN }

Based on the segmentation bounding boxes, we applied our Latent CNN (LCNN) framework, and achieved the state-of-the-art performance: mean average precision of 81.4\% as shown in Table~\ref{tab:Cls07}.

\begin{table*}[t]
  \caption{Pascal VOC 2007 Image Classification Results(average precision \%)}
  \label{tab:Cls07}
  \vskip 0.15in
  \begin{center}
  \begin{small}
  \begin{sc}
 \setlength{\tabcolsep}{3pt}
 \begin{tabular}{|c||c|c|c|c|c|c|c|c|c|c|c|}
\hline
\;\; &plane&bike&bird& boat& bottle& bus& car& cat& chair& cow
\\\hline
SuperVec~\cite{SuperVec} &79.4&72.5&55.6&73.8&34.0&72.4&83.4&63.6&56.6&52.8
\\\hline
GHM~\cite{GHM}&76.7&74.7&53.8&72.1&40.4&71.7&83.6&66.5&52.5&57.5
\\\hline
NUS~\cite{song2011contextualizing} &82.5&79.6&64.8&73.4&54.2&75.0&87.5&65.6&62.9&56.4
\\\hline
AGS~\cite{AGS} &82.2&83.0&58.4&76.1&56.4&77.5&88.8&69.1&62.2&61.8
\\\hline
CNNSVM~\cite{offtheshelf} &88.5&81.0&83.5&82.0&42.0&72.5&85.3&81.6&59.9&58.5
\\\hline
CNNaugSVM ~\cite{offtheshelf} &\bf90.1&84.4&86.5&84.1&48.4&73.4&86.7&85.4&61.3&67.6
\\\hline
Patch-seg &86.9&\bf85.3&87.7&82.9&60.4&79.3&89.8&86.1&63.7&75.8
\\\hline
LCNN-seg &88.9&84.2&\bf88.1&\bf84.4&\bf62.6&\bf80.0&\bf90.2&\bf87.0&\bf65.7&\bf76.7
\\\hline
\;\; &table& dog& horse& motor& person& plant& sheep& sofa& train& tv& mAP
\\\hline
SuperVec~\cite{SuperVec} &63.2&49.5&80.9&71.9&85.1&36.4&46.5&59.8&83.3&58.9&64.0
\\\hline
GHM~\cite{GHM}&62.8&51.1&81.4&71.5&86.5&36.4&55.3&60.6&80.6&57.8&64.7
\\\hline
NUS~\cite{song2011contextualizing} &66.0&53.5&85.0&76.8&91.1&53.9&61.0&67.5&83.6&70.6&70.5
\\\hline
AGS~\cite{AGS} &64.2&51.3&\bf85.4&80.2&91.1&48.1&61.7&67.7&86.3&70.9&71.1
\\\hline
CNNSVM~\cite{offtheshelf} &66.5&77.8&81.8&78.8&90.2&54.8&71.1&62.6&87.2&71.8&73.9
\\\hline
CNNaugSVM ~\cite{offtheshelf} &69.6&84.0&\bf85.4&80.0&92.0&56.9&76.7&67.3&89.1&74.9&77.2
\\\hline
Patch-seg &73.4&\bf85.6&83.6&84.6&\bf94.1&68.8&80.1&\bf74.1&\bf89.6&79.8&80.6
\\\hline
LCNN-seg &\bf73.8&85.3&85.1&\bf86.5&93.8&\bf69.5&\bf83.8&73.6&89.2&\bf80.4&\bf81.4
\\\hline

\end{tabular}
\end{sc}
\end{small}
\end{center}
\vskip -0.1in
\end{table*}


\subsection{CIFAR-10}
The CIFAR-10 dataset consists of 60,000 32x32 color images in 10 classes, with 6,000 images per class. There are 50,000 training images and 10,000 test images. The dataset is divided into five training batches and one test batch, each with 10,000 images. The test batch contains exactly 1,000 randomly-selected images from each class. The training batches contain the remaining images in random order.

\begin{table}[h]
  \caption{Test set error rates for CIFAR-10 of various methods}
  \label{tab:cifar10}
  \vskip 0.15in
  \begin{center}
  \begin{small}
  \begin{sc}
  \begin{tabular}{p{6cm} c}
    \hline
    Method                                                     & Test Error \\
    \hline
    Stochastic Pooling~\cite{StoPool}                          & 15.13\% \\
    CNN + Spearmint~\cite{Spearmint}                           & 14.98\% \\
    Conv. maxout + Dropout ~\cite{Maxout}                      & 11.68\% \\
    NIN + Dropout~\cite{NIN}                                   & 10.41\% \\
    CNN + Spearmint + Data Augmentation~\cite{Spearmint}       &  9.50\% \\
    Conv. maxout + Dropout + Data Augmentation~\cite{Maxout}   &  9.38\% \\
    DropConnect + 12 networks + Data Augmentation~\cite{Dropcon} &  9.32\% \\
    NIN + Dropout + Data Augmentation~\cite{NIN}               &  8.81\% \\
    \hline
    \hline
    global CNN                                                       & 11.3\% \\
    part   CNN                                                         & 10.7\% \\
    2CNN-average                                                       & 9.4\% \\
    2CNN-latent                                                        & 8.13\% \\
    
  \end{tabular}
  \end{sc}
  \end{small}
  \end{center}
  \vskip -0.1in
\end{table}

The deep features are extracted by NIN~\cite{NIN}, which is stacked by three mlpconv layers, followed by a pooling layer and a dropout layer, and then the final global average pooling layer. The parameters in first mlpconv layer are (3x5x5)x192, 192x160x96. 3x5x5 is the receptive field for the input image, which is also the filter size for classical convolutional layers, while 192 is the number of such filters. 192x160x96 is the size for the mlp layer. The pooling layer is 3x3 max pooling with stride 2. Then a dropout layer is added to the mlpconv, because a three-layer tends to overfit.  The parameters in second mlpconv layer are (96x5x5)x192, 192x192x192. Also, it is followed by 3x3 max pooling with stride 2 and dropout layer. The parameters in the third mplconv layer are (192x3x3)x192, 192x192x10, which is also followed by 3x3 max pooling with stride 2 and dropout layer. The average global pooling and softmax layer are added to form the structure of the NIN network.

Two CNNs are trained separately with whole image (32x32x3) and image patches (24x24x3), but those two CNNs share the same structure setup as above. For the latent SVM fusion part, 10 locations are extracted for the part CNN and the response from the best location is combined with the response of the whole image, as shown in Algorithm \ref{algm:MLSVM2}.

From the Table~\ref{tab:cifar10}, we achieved 8.13\% error rate and outperform the NIN by 0.68 percent.

\subsection{CIFAR-100}
This dataset is just like CIFAR-10, except it has 100 classes containing 600 images each. There are 500 training images and 100 testing images per class. 

The NIN for CIFAR-100 is almost the same as the NIN for CIFAR-10 except that the parameter for the thrid mlp is  (192x3x3)x192, 192x192x100, because the category number is 100.

From the Table~\ref{tab:cifar100}, we achieve 33.73\% error rate and outperform NIN by almost 2 percents.

\begin{table}[h!]
  \caption{Test set error rates for CIFAR-100 of various methods}
  \label{tab:cifar100}
  \vskip 0.15in
  \begin{center}
  \begin{small}
  \begin{sc}
  \begin{tabular}{p{6cm} c}
    Method                                                     & Test Error \\
    \hline
    Learned Pooling~\cite{LearnedPool}                          & 43.71\% \\
    Stochastic Pooling~\cite{StoPool}                          & 42.51\% \\
    Conv. maxout + Dropout ~\cite{Maxout}                      & 38.57\% \\
    Tree based priors~\cite{TreePrior}                         & 36.85\% \\
    NIN + Dropout~\cite{NIN}                                   & 35.68\% \\
    \hline
    global CNN                                                  & 36.44\% \\
    part   CNN                                                 & 34.72\% \\
    2CNN-average                                               & 33.34\% \\
    2CNN-latent                                                & 32.31\% \\
  \end{tabular}
  \end{sc}
  \end{small}
  \end{center}
  \vskip -0.1in
\end{table}

\subsection{MNIST}
The MNIST database of handwritten digits, has a training set of 60,000 examples and a test set of 10,000 examples. 

The deep features are extracted by NIN~\cite{NIN}, which is stacked by three mlpconv layers, followed by a pooling layer and a dropout layer, and the final global average pooling layer. The parameters in first mlpconv layer are (1x5x5)x96 and 96x64x48. 1x5x5 is the receptive field for the input image; 1 means that it is gray images, while 96 is the number of such filters. 96x64x48 is the size for the mlp layer. The pooling layer is 3x3 max pooling with stride 2. Then a dropout layer is added to the mlpconv, because a three-layer tends to overfitting.  The parameters in second mlpconv layer are (48x5x5)x128, 128x96x48. Also, it is followed by 3x3 max pooling with stride 2 and dropout layer. The parameters in third mplconv layer is (48x3x3)x128, 128x96x10, which is also followed by 3x3 max pooling with stride 2 and dropout layer. The average global pooling and softmax layer are added to form the structure of the NIN network.

Two CNNs are trained separately with whole image (28x28x1) and image patches (24x24x1), but those two CNNs share same structure setup as above. For the latent SVM fusion part, 10 locations are extracted for the part CNN and the response from the best location is combined with the response of the whole image, as Algorithm \ref{algm:MLSVM2}. From the Table~\ref{tab:mnist}, we achieve 0.42\% error rate and outperform the NIN by 0.05 percent.

\begin{table}[ht]
  \caption{Test set error rates for MNIST of various methods}
  \label{tab:mnist}
  \vskip 0.15in
  \begin{center}
  \begin{small}
  \begin{sc}
  \begin{tabular}{p{6cm} c}
    Method                                                     & Test Error \\
    \hline
    2-Layer CNN + 2-Layer NN~\cite{StoPool}                    & 0.53\% \\
    Stochastic Pooling~\cite{StoPool}                          & 0.47\% \\
    NIN + Dropout~\cite{NIN}                                   & 0.47\% \\
    Conv. maxout + Dropout ~\cite{Maxout}                      & 0.45\% \\
    \hline
    global CNN                                                 & 0.50\% \\
    part   CNN                                                 & 0.81\% \\
    2CNN-average                                               & 0.48\% \\
    2CNN-latent                                                & 0.42\% \\
  \end{tabular}
\end{sc}
\end{small}
\end{center}
\vskip -0.1in
\end{table}


\section{Conclusion}

We introduced a novel multiple CNN combination method by latent SVM. We also developed the stochastic gradient descent version of multi-class latent SVM. The experiments show that this technique works well with both AlexNet and NIN for classification tasks.

\section*{Acknowledgment}
The authors would like to thank Sony Electronics Inc. for their generous funding of this work.  

\IEEEtriggeratref{7}

\bibliographystyle{IEEEtran}
\bibliography{root}

\end{document}